\documentclass[runningheads]{llncs}
\usepackage{graphicx}
\usepackage{graphicx}
\usepackage{amsmath}
\usepackage{amssymb}
\usepackage{booktabs}
\usepackage{algorithm}
\usepackage{algorithmic}
\usepackage{multirow}
\usepackage{url}
\usepackage{hyperref}
\begin{document}
\title{Multi-Modal Dialogue State Tracking for Playing GuessWhich Game}

%
\author{Wei Pang\inst{1} \and
Ruixue Duan\inst{1} \and
Jinfu Yang\inst{2} \and 
Ning Li\inst{1}}
\authorrunning{Wei Pang et al.}
%
\institute{Beijing Information Science and Technology University, Beijing, China
\email{\{pangweitf,duanruixue\}@bistu.edu.cn}\\
\and
Beijing University of Technology, Beijing, China\\}
\maketitle              
\begin{abstract}
GuessWhich is an engaging visual dialogue game that involves interaction between a Questioner Bot (QBot) and an Answer Bot (ABot) in the context of image-guessing. In this game, QBot's objective is to locate a concealed image solely through a series of visually related questions posed to ABot. However, effectively modeling visually related reasoning in QBot's decision-making process poses a significant challenge. Current approaches either lack visual information or rely on a single real image sampled at each round as decoding context, both of which are inadequate for visual reasoning. To address this limitation, we propose a novel approach that focuses on visually related reasoning through the use of a mental model of the undisclosed image. Within this framework, QBot learns to represent mental imagery, enabling robust visual reasoning by tracking the dialogue state. The dialogue state comprises a collection of representations of mental imagery, as well as representations of the entities involved in the conversation. At each round, QBot engages in visually related reasoning using the dialogue state to construct an internal representation, generate relevant questions, and update both the dialogue state and internal representation upon receiving an answer. Our experimental results on the VisDial datasets (v0.5, 0.9, and 1.0) demonstrate the effectiveness of our proposed model, as it achieves new state-of-the-art performance across all metrics and datasets, surpassing previous state-of-the-art models. Codes and datasets from our experiments are freely available at \href{https://github.com/xubuvd/GuessWhich}.

\keywords{GuessWhich \and Multi-Modal Dialogue State Tracking \and Visual Dialogue.}
\end{abstract}
\section{Introduction}

\begin{figure}[!htb]
\centering
\includegraphics[width=0.98\textwidth,scale=0.8, clip=true, keepaspectratio]{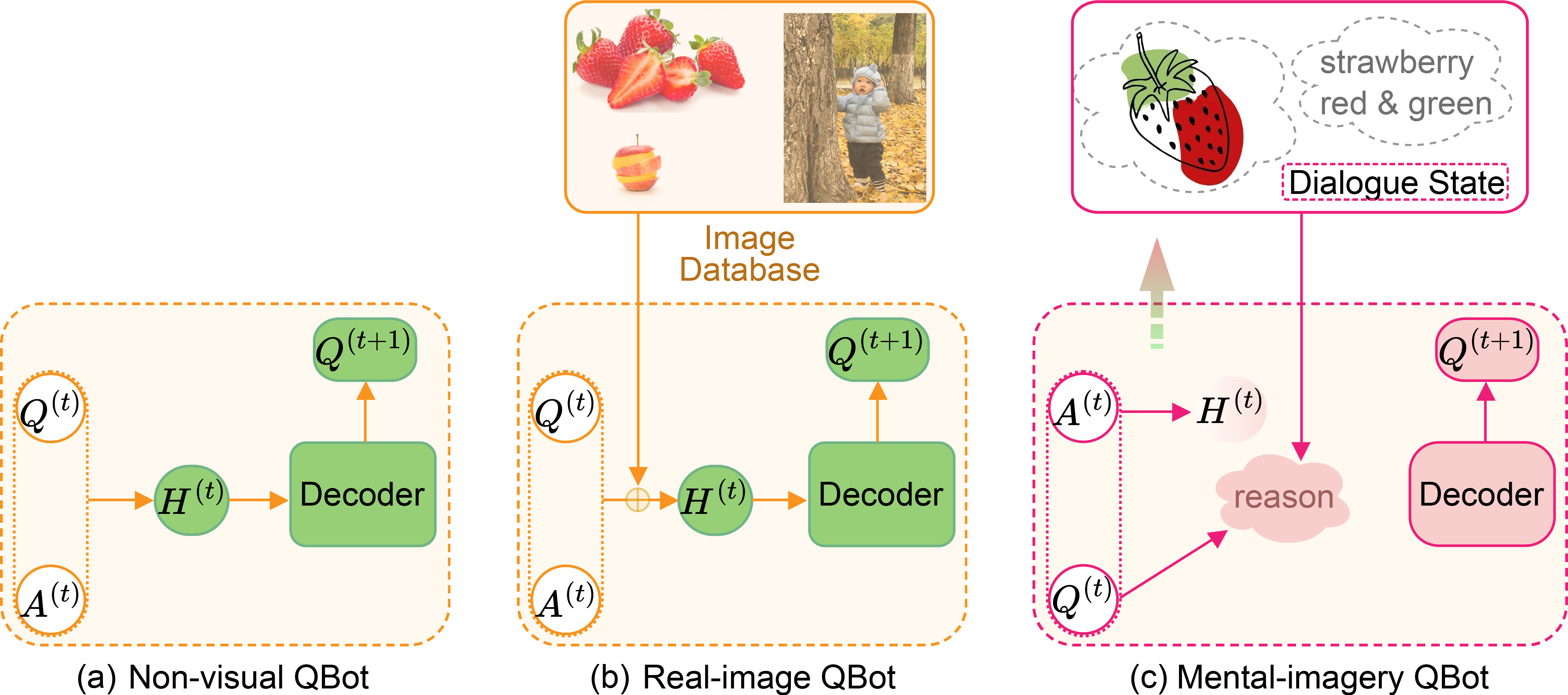}
\caption{Illustration of three types of QBot involving four components: question (Q), answer (A), history encoder (H) at round t, and Decoder for new question generation. Specially, a) Non-visual QBot, no visual information is provided to QBot. b) Real-image QBot, retrieves a real image per round from a pool to serve as visual information for Decoder. c) Mental-imagery QBot (Ours), explores visually related reasoning based on the QBot's mental model of the secret image.}
\label{fig_idea}
\end{figure}

In the future, visual conversational agents may engage in natural language conversations. Research in vision-language tasks is still in its early stages, progressing from single-round VQA \cite{VQA}  to multi-round Visual Dialogue and multi-modal multi-modal Video Dialogue \cite{VD,GuessWhat,Game,Image-Chat}. Among these tasks, GuessWhich stands out as a two-player image-guessing game with a QBot and ABot. QBot aims to identify a hidden image by asking questions to ABot. While ABot has received attention, research on QBot is relatively limited, which is the focus of this paper.

QBot starts by receiving a caption and formulating the first question. Subsequent questions are generated based on the caption and dialogue history. Performing visual reasoning solely from textual information poses a challenge. Existing approaches, as illustrated in Fig.\ref{fig_idea} (a) and (b), can be divided into non-visual methods \cite{Game,RL,AQM,ADQ,Devil,ReeQ,AMN} that rely on language-based models and real-image methods \cite{ImgGue} that retrieve probable images as visual context. However, both approaches have limitations in visual reasoning.

Firstly, QBot needs to generate questions that pertain to the image, linking words to visual concepts, much like humans do \cite{Paivio}. Hence, QBot, which lacks visual information in its modeling, is inadequate for this purpose. Secondly, QBot samples a single real image from a large pool of images, often numbering in the thousands, e.g., 9,628 candidate images per round. This approach is unnatural in the context of a genuine game and introduces substantial sampling variance, rendering the model's reasoning process unstable.

Drawing on the Dual-coding theory \cite{Paivio}, human cognition is based on two interconnected modes: words and images. Mental imagery plays a significant role in word comprehension. Building on this theory, we propose aligning dialogue entities with mental visual concepts in QBot, as depicted in Fig.\ref{fig_idea}(c). By constructing a mental model of the secret image through textual semantics, we establish a dialogue state consisting of representations of dialogue entities and mental objects denoted as $\langle$words, images$\rangle$ states.

As the game progresses, the dialogue state evolves, encompassing representations of mental objects in QBot's mind, and prompting QBot to pose visually related questions. To the best of our knowledge, the problem of modeling visual reasoning in QBot remains relatively unexplored. In this paper, we present a QBot model that incorporates mental imagery through dialogue state tracking (DST) to address the aforementioned ideas. Our model consists of a cycle of two primary procedures: Visual Reasoning on Dialogue State (VRDS) and State Tracking (STrack).

VRDS facilitates three-hop reasoning in the dialogue state, progressing through the path $\mathrm{words}{\rightarrow} \mathrm{words}{\rightarrow} \mathrm{images}{\rightarrow} \mathrm{words}$ and generating an internal representation. A decoder utilizes this representation to generate new questions. Upon receiving an answer, STrack is activated, involving two actions: 1) Addition, introducing new textual semantics to the dialogue states (e.g., "strawberry" as a new entity and its associated mental object). 2) Update, incorporating new textual features (e.g., colors, positions, counts) into the aligned dialogue states' existing representation. Experimental results demonstrate our model's superior performance, achieving a new state-of-the-art level. In summary, our contributions are three-fold:
\begin{itemize}
\item We propose a novel QBot agent that is capable of performing visually related reasoning based on mental imagery in one's mind in dialog.

\item We present dialogue state tracking based QBot model (DST), which learns to form representations of mental imagery that support visually related reasoning. The dialogue states, composed of not only words states but also images states, are tracked and updated through dialoguing with ABot.

\item Achieving new state-of-the-art results on the GuessWhich game underlying VisDial v0.5, v0.9 and v1.0 datasets. Compared with prior studies, this work takes a step towards mimicking humans playing a series of visual dialogue games (such as GuessWhich).
\end{itemize}

\section{Related Work}
Visual Dialogue is a key area of research in vision-language studies, with the aim of developing conversational agents capable of human-like interactions. Recent progress \cite{MRDST,VDST,GST} has been made in various tasks, including GuessWhat!?, GuessWhich, and Visual \& Video Dialogue. GuessWhich specifically involves the challenge of QBot finding an undisclosed image from a large pool without sharing it with ABot.

Existing QBot models can be categorized into non-visual and real-image approaches. Non-visual models, represented in Fig.\ref{fig_idea}(a), do not utilize visual information. For example, Das et al. \cite{RL} propose an encoder-decoder network with a feature regression network. They use a hierarchical encoder-decoder architecture to generate questions based on history. Murahari et al. \cite{ADQ} introduce a Smooth-L1 Penalty to mitigate repetitive questions. Zhao et al. \cite{AMN} incorporate an Attentive Memory Network, and Li et al. \cite{AQM} propose an information-theoretic model. These models solely rely on textual information.

Real-image models, shown in Fig.\ref{fig_idea}(b), provide physical images as input to the QBot decoder. Zhou et al. \cite{ImgGue} introduce an image-guesser module into the QBot model. While related works track real image objects, our approach focuses on constructing and tracking mental imagery representations during the dialogue.

\section{Model}
\begin{figure*}[!htb]
\centering
\includegraphics[width=0.99\textwidth,scale=0.5, clip=true,keepaspectratio]{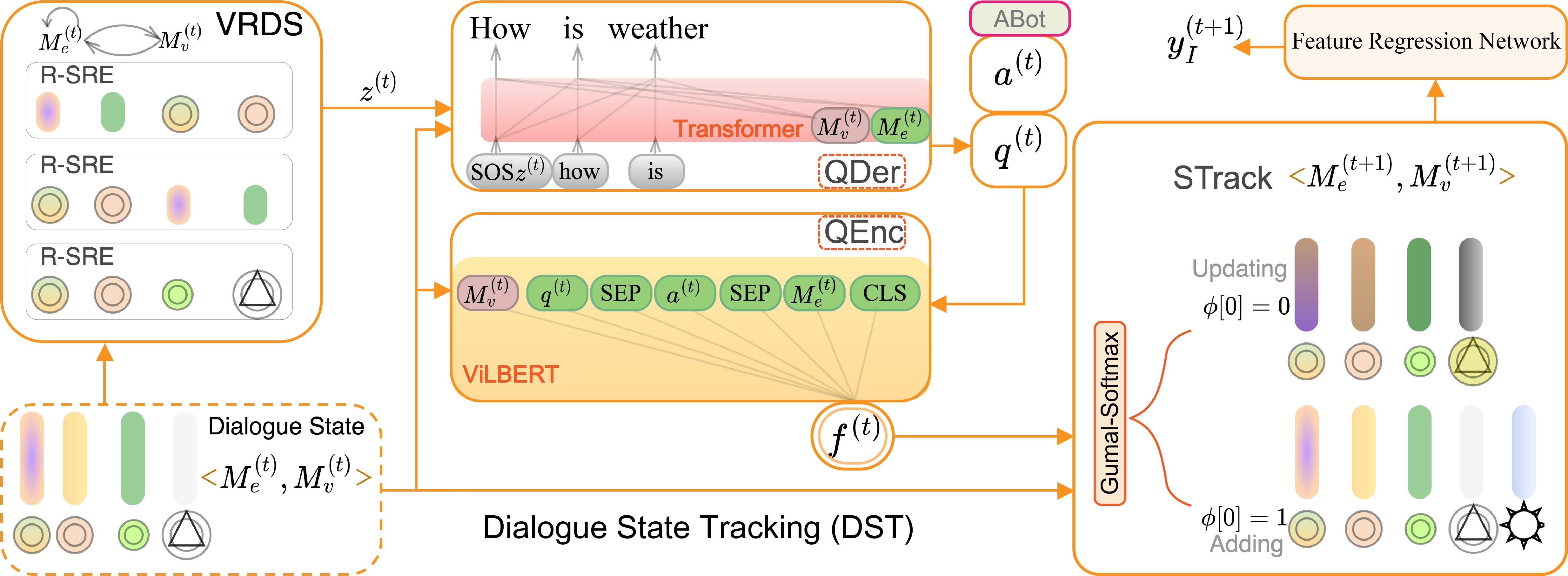}
\caption{Overall structure of the proposed DST model. The oblong colorful strips denote image state $M^{(t)}_{v}$, and the rounded circle are word state $M^{(t)}_{e}$.}
\label{fig_overview}
\end{figure*}

The proposed Dialogue State Tracking based QBot (DST) model, depicted in Fig.\ref{fig_overview}, consists of five modules: Recursive Self-Reference Equation (R-SRE), Visual Reasoning on Dialogue State (VRDS), Question Decoder (QDer), QBot Encoders (QEnc), and State Tracking (STrack). Detailed explanations of each module will be provided in the subsequent sections.

\noindent {\bf Problem Setting}
At the start of the game, QBot receives a caption $C$ describing the target image $I^{*}$ visible to ABot. This caption provides the initial textual information for QBot to generate its first question. In subsequent rounds, QBot generates questions $q^{(t)}$ based on the accumulated dialogue history $H$, which includes the caption $C$ and previous question-answer pairs. QBot's goal is to use this information to guess the undisclosed image from a candidate pool.

We define an accumulated dialogue state ${<}M^{(t)}_{e}, M^{(t)}_{v}{>}$ at round $t$. The words state $M^{(t)}_{e}$ represents the textual representation of discussed entities, while the images state $M^{(t)}_{v}$ represents mental imagery information derived from the words in QBot's mind. The initial dialogue state ${<}M^{(1)}_{e}, M^{(1)}_{v}{>}$ is constructed using the caption $C$ as input and the Adding action of the STrack module. Both $M^{(1)}_{e}$ and $M^{(1)}_{v}$ are $1\times d$ matrices.

\noindent {\bf Recursive Self{-}Reference Equation (R{-}SRE)}
To capture visually related interactions within and between modalities in the dialogue state, we propose a recursive self-reference equation (R-SRE) denoted as $V' = \mathrm{R-SRE}(Q, V)$. The R-SRE consists of two attention mechanisms that facilitate the update of the matrix $V$ based on the guidance provided by another matrix $Q$. The equation is formulated as follows:
\begin{align}
\label{equ_sre1}\alpha &= \mathrm{softmax}(w_{\alpha}Q),  \\
\label{equ_sre2}q &= \alpha^{T} Q, \\
\label{equ_sre3}\beta &= \mathrm{softmax}(w_{\beta}[\mathrm{r}^{(k)}(q); \mathrm{r}^{(k)}(q)\odot V;V]), \\
\label{equ_sre4}V^{\prime} &= \beta V,
\end{align}
where $Q, V\in \mathbb{R}^{k\times d}$ are two input data matrix, $w_{\alpha}\in \mathbb{R}^{d\times 1}$ and $w_{\beta}\in \mathbb{R}^{3d\times 1}$ are trainable projection matrices. [;;] denotes the symmetric concatenation between bi-modalities as in \cite{GST,VDST}, $\odot$ is the element-wise product. We define a repeat operation $\mathrm{r}^{(k)}(q)$ that repeats $q$ k times to form a matrix of size $k\times d$ with same dimension as $V$. Specially, a self-attention on $Q$ is first performed and attention scores $\alpha\in \mathbb{R}^{k\times 1}$ is obtained, $q\in \mathbb{R}^{d\times 1}$ is the weighted sum of $Q$ using $\alpha$. Then, $q$ is referred to as a key again to query $V$ to get a $\beta\in \mathbb{R}^{k\times 1}$ over $V$. Finally, $V$ is changed by multiplying the attention weight $\beta$ to yield a new same-dimension representation $V^{\prime}$. We omit bias where possible, for clarity.

\noindent {\bf Visual Reasoning on Dialogue State (VRDS)}

The VRDS process involves three-hops reasoning using the R-SRE operation. Firstly, a self R-SRE operation updates the words state (Eq.\ref{equ_VRDS1}). Then, a cross-modal R-SRE operation generates a visually related context vector (Eq.\ref{equ_VRDS2}). Iterative cross-modal R-SRE operations between the updated words state and the images state yield an intermediate textual context vector (Eq.\ref{equ_VRDS3}). Finally, the visually and textually related context vectors are concatenated and processed by a linear layer to obtain the final context vector. Formally,
\begin{align}
\label{equ_VRDS1} \widehat{M}^{(t)}_{e} &= \mathrm{R{-}SRE}(M^{(t)}_{e}, M^{(t)}_{e}),\\
\label{equ_VRDS2} \widehat{M}^{(t)}_{v} &= \mathrm{R{-}SRE}(\widehat{M}^{(t)}_{e},M^{(t)}_{v}), a_{v}^{(t)} = \mathrm{sum}(\widehat{M}^{(t)}_{v}),\\
\label{equ_VRDS3} \widetilde{M}^{(t)}_{e} &= \mathrm{R{-}SRE}(\widehat{M}^{(t)}_{v}, \widehat{M}^{(t)}_{e}), a_{e}^{(t)} = \mathrm{sum}(\widetilde{M}^{(t)}_{e}),\\
\label{equ_VRDS4} z^{(t)} &= \mathrm{Dropout}(w_{v}[a_{v}^{(t)}; a_{e}^{(t)}]),
\end{align}
where $w_{v}\in \mathbb{R}^{2d\times d}$ are learnable projection matrixes, $sum$ is operated on the k-dimension to compact the intermediate representation matrix to a vector, $z^{(t)}\in \mathbb{R}^{d}$ is the final context vector passed to QDer module.

From Eq.\ref{equ_VRDS1} to Eq.\ref{equ_VRDS4}, VRDS performs textually and visually related interactions upon dialogue state. It models the cross{-}modal interactions between two states motivated from Dual-coding theory\cite{Paivio}: QBot can think with words as well as can think with images based on mental imagery representations in one's mind.

\noindent {\bf Question Decoder (QDer)}
We use a multi-layer Transformer decoder \cite{Transformer} and employs softmax on its output. It takes $z^{(t)}$ and dialogue states as input. The QDer module predicts next word by employing cross-attention with dialogue states as conditioning, continuing this process until an end-of-sequence token [EOS] is encountered. Formally,
\begin{align}\label{equ_QDer}
h^{(t)}_{i+1} &= \mathrm{transformer\_decoder}(h^{(t)}_{i}, M^{(t)}_{e}, M^{(t)}_{v}),\\
w^{(t)}_{i+1} &= \mathit{argmax}\ \mathrm{softmax}(h^{(t)}_{i+1}w_{der}),
\end{align}
where $w_{der}\in \mathbb{R}^{d\times |\mathrm{V}|}$ is trainable parameters, $\mathrm{V}$ is the number of vocabulary. $h^{(t)}_{i+1}$ denotes the representation of next token $w^{(t)}_{i+1}$, which is selected by the greedy algorithm. $h^{(t)}_{0}$ is initialized as the element-wise addition of the starting token [SOS] and $z^{(t)}$. We save a sequence of $w^{(t)}_{i+1}$ to produce a new question $q^{(t)}$.

\noindent {\bf QBot Encoders (QEnc)}
QEnc utilizes a pre-trained vision-linguistic model called ViLBERT \cite{ViLBERT}. The input to ViLBERT is structured as follows: [CLS] $q^{(t)}$ [$\mathrm{SEP}$] $a^{(t)}$ [$\mathrm{SEP}$] $M^{(t)}{e}$ [$\mathrm{SEP}$] $M^{(t)}{v}$, with each segment separated by a [SEP] token. The output of [CLS] token is considered as the fact representation $f^{(t)}\in \mathbb{R}^{d}$, which captures the information from new question-answer pair.

\noindent {\bf State Tracking (STrack)}
STrack offers two actions: Adding and Updating. The decision between these actions is determined by a differentiable binary choice made using Gumbel-Softmax sampling \cite{gumbel}. This allows for end-to-end training. We introduce a two-layer feedforward network denoted as $\mathrm{FFN}(\cdot)$, which incorporates GELU activation, Dropout, and LayerNorm. For clarity, we apply $\mathrm{FFN}(\cdot)$ to the element-wise product of the fact representation $f^{(t)}$ and words state $M^{(t)}_{e}$. Subsequently, a Gumbel-Softmax operation is performed, yielding a probability distribution representing the action type.

\begin{equation} \label{equ5}
\begin{split}
\phi &= \mathrm{Gumbel}(\mathrm{FFN}([\mathrm{r}^{(k)}(f^{(t)}) \odot M^{(t)}_{e}])),\\
\end{split}
\end{equation}
where $\phi\in \mathbb{R}^{2}$ is 2-d one-hot vector for discrete decision. According to predicted $\phi$, one of the two actions is chosen for the STrack.

\noindent {\bf Adding Action on Words State}
The Adding action is executed on the current words state if $\phi[0]=1$. It takes the $f^{(t)}$ as input to $\mathrm{FFN}$, resulting in a new textual representation $e^{(t+1)}{a,w}\in \mathbb{R}^{d}$. Subsequently, $e^{(t+1)}{a,w}$ is inserted into the set of words states, leading to a new set of words states $M^{(t+1)}_{e}\in \mathbb{R}^{(k+1)\times d}$:
\begin{align}
\label{equ_aaws1} e^{(t+1)}_{a,w} &= \mathrm{FFN}(f^{(t)}),\\
\label{equ_aaws2} M^{(t+1)}_{e} &= M^{(t)}_{e} \cup \{e^{(t+1)}_{a,w}\},
\end{align}
where $\cup$ is an append operation. Note that the size of newly updated $M^{(t+1)}_{e}$ is increased to $k+1$.

\noindent {\bf Adding Action on Images State}
In the case of $\phi[0]=1$, the adding action is performed on current images states. It produces a new mental object $o^{(t+1)}_{a,v}\in \mathbb{R}^{d}$, which denotes a new visual concept (e.g., “carpet”). We translate the fact representation into images with $\mathrm{FFN}$ network, and get a new set of images states $M^{(t+1)}_{v}\in \mathbb{R}^{(k+1)\times d}$, in which its size is increased to $k+1$. Formally,
\begin{align}
\label{equ_aais1} o^{(t)}_{a,v} &= \mathrm{FFN}(f^{(t)}),\\
\label{equ_aais2} M^{(t+1)}_{v} &= M^{(t)}_{v} \cup \{o^{(t)}_{a,v}\},
\end{align}
where fact representation $f^{(t)}$ is used for translation from words to images by $\mathrm{FFN}$.

\noindent {\bf Updating Action on Words State}
When $\phi[0]=0$, the updating action is applied to the words state. It includes calculating an assignment distribution $\psi$ that determines how much the new fact representation can be merged into the existing representation of the associated words state. This is achieved by passing $f^{(t)}$ and the current words state $M_{e}^{(t)}$ through a two-layer feedforward network and a softmax classifier.

\begin{align}
\label{equ_uaws1} \psi &= \mathrm{softmax}(\mathrm{FFN}([\mathrm{r}^{(k)}(f^{(t)})\odot M_{e}^{(t)}])),\\
\label{equ_uaws2} M^{(t+1)}_{e}&= M^{(t)}_{e} + \psi \mathrm{FFN}_{\psi}(f^{(t)}),
\end{align}
where $\psi\in \mathbb{R}^{k}$ is the assignment distribution, $\mathrm{FFN}_{\psi}(\cdot)\in \mathbb{R}^{d}$ denotes another $\mathrm{FFN}$ network, $M^{(t+1)}_{e}\in \mathbb{R}^{k\times d} $ is the newly updated words states. Note that the number of words states in $M^{(t+1)}_{e}$ remains unchanged in this case.

\noindent  {\bf Updating Action on Images State}
If $\phi[0]=0$, similar to the case of updating on words state, we compute an assignment distribution for associating current fact representation with previous images state. Formally,
\begin{align}
\label{equ_uais1} \gamma &= \mathrm{softmax}(\mathrm{FFN}([\mathrm{r}^{(k)}(f^{(t)})\odot M_{v}^{(t)}])),\\
\label{equ_uais2} M^{(t+1)}_{v} &= M^{(t)}_{v} + \gamma \mathrm{FFN}_{\gamma}(f^{(t)}),
\end{align}
where $\gamma\in \mathbb{R}^{k}$ is the assignment distribution, $\mathrm{FFN}_{\gamma}(\cdot)\in \mathbb{R}^{d}$ is another $\mathrm{FFN}$ network different from in Eq.\ref{equ_uais1}. It converts fact representation to visually related representation, which accumulates further visual attributes (such as object shape, color and position, e.g., a question “what color is the carpet?” with an answer “red”) about the contents of the undisclosed image in the same representation in $M_{v}^{(t)}$. $M^{(t+1)}_{v}\in \mathbb{R}^{k\times d}$ is the newly updated images state, its size remains unchanged.

\noindent {\bf Model Training}
Our model is optimized using supervised learning (SL) with three loss functions: Cross-Entropy (CE) loss, Mean Square Error (MSE) loss, and Progressive (PL) loss. The CE loss is computed based on ground truth questions, while the MSE loss is calculated using the feature regression network $f(\cdot)$ \cite{RL} to predict the image representation $y^{(t)}{I}$. The MSE loss compares the predicted representation $y^{(t)}{I}$ with the ground truth image representation $y^{(t)}_{I^{*}}$ obtained from VGG19.
\begin{equation} \label{equ10}
\mathcal{L}_{\mathrm{CE}} = -\frac{1}{l}\sum_{j=1}^{l}\log p_{j}; \mathcal{L}_{\mathrm{MSE}} = -\frac{1}{T}\sum_{t=1}^{T}||y^{(t)}_{I^{*}} - y^{(t)}_{I}||^{2}_{2},
\end{equation}
\noindent where $l$ here denotes the total length of generated questions, T is the total dialog rounds, $p_{j}$ is the probability of ground-truth word at step j in the dialogue.

Because of multi-round dialogue brings a series of MSE loss, we present a progressive loss that is defined as the difference of MSE loss in successive dialog rounds, which encourages Questioner to progressively increase similarity towards the target image, as written in: $\mathcal{L}_{\mathrm{PL}} = -\frac{1}{T-1}\sum_{t=2}^{T}\mathcal{L}^{(t)}_{\mathrm{MSE}}-\mathcal{L}^{(t-1)}_{\mathrm{MSE}}$. Overall, the final loss for supervised learning QBot is a sum of three losses as mentioned above: $\mathcal{L}_{SL} = \mathcal{L}_{\mathrm{CE}} + \mathcal{L}_{\mathrm{MSE}} + \mathcal{L}_{\mathrm{PL}}$.

\section{Experiment and Evaluation}

\noindent {\bf Dataset}
Our GuessWhich model is evaluated on three benchmarks: VisDial v0.5, v0.9, and v1.0. These datasets include various numbers of training, validation, and test images. VisDial v1.0 has 123,287 training images, 2,064 validation images, and 8,000 test images. VisDial v0.9 includes 82,783 training images and 40,504 validation images. VisDial v0.5 consists of 50,729 training images, 7,663 validation images, and 9,628 test images. Dialogues in these datasets contain a caption for the target image and multiple question-answer pairs. It's worth noting that only the test set of VisDial v1.0 has variable-length dialogues, while the other dataset splits have fixed 10-round dialogues.

\noindent {\bf Evaluation Metric}
We follow the standard evaluation metrics \cite{RL,ADQ} for QBot in two parts: image guessing and question diversity. At image guessing, we report retrieval metrics of target image, including mean reciprocal rank (MRR), Recall @k (R@k) for k = {1, 5, 10}, mean rank (Mean) and percentile mean rank (PMR). At question diversity, we adopt six metrics like Novel Questions \cite{ADQ}, Unique Questions \cite{ADQ}, Dist-n and Ent-n \cite{Dist-n}, Negative log-likelihood \cite{ADQ}, and Mutual Overlap \cite{Overlap}.

\noindent {\bf Implementation Details}
Our model architecture consists of a cross-modal Transformer decoder with 12 layers and a hidden state size of 768. It utilizes 12 attention heads. The base encoder is a pre-trained ViLBERT model with 12 layers and a hidden state size of 768. During training, we used a batch size of 64 and trained the model for 30 epochs. A dropout rate of 0.1 was applied after each linear layer. Early stopping was implemented on the validation split if the performance metric (PMR) did not improve for 10 consecutive epochs. We used the Adam optimizer with a base learning rate of 1e-3, which decayed to 1e-5 during training. For image representation, we used pre-extracted VGG19 features, where each image is represented by a 4096-dimensional vector.

\noindent {\bf Comparison to State-of-the-Art Methods}
The comparing methods on QBot can be regarded to have three types: 1) Non-visual based models, like SL{-}Q \cite{RL}, ReCap\cite{Devil}, ADQ\cite{ADQ}, AMN\cite{AMN}, RL{-}Q \cite{RL}, AQM{+}{/}indA\cite{AQM}, ReeQ-SL (trained in SL) \cite{ReeQ} and ReeQ-RL (fine-tuned in reinforcement learning)\cite{ReeQ}. 2) Real-image based methods, such as SL{-}Q{-}IG \cite{ImgGue}. and 3) Mental-imagery based method: our DST. 

\begin{table*}[!htb]
\centering
\caption{Result comparison of image guessing on VisDial datasets. Higher is better for MRR, R@k, and PMR, and lower is better for Mean. Note that $^{\dagger}$ means we roughly estimated the value of Mean by an approximate evaluation: $\mathrm{Mean} \simeq \mathrm{\#Num\ of\ Image\ Pool}\times(1.0 - \mathrm{PMR})$\cite{Devil}, and the results of $^{\diamond}$ are cited from \cite{AMN}.}\label{tab_guess}
\setlength{\tabcolsep}{0.05mm}{
\begin{tabular}{l|cccccc|l|c}
\hline
Model& MRR$\uparrow$ & R@1$\uparrow$ & R@5$\uparrow$ & R@10$\uparrow$ & PMR$\uparrow$ & Mean$\downarrow$ & Dataset&\#Num of Image Pool\\
\hline
SL{-}Q\cite{RL}&-&-&-&-&91.19&848.2$^{\dagger}$&\multirow{6}{*}{v0.5 test}&\multirow{6}{*}{9,628}\\
RL{-}Q\cite{RL}&-&-&-&-&94.19&559.4$^{\dagger}$&&\\
SL{-}Q{-}IG\cite{ImgGue}&-&-&-&-&96.09&376.5$^{\dagger}$&&\\
ReCap\cite{Devil}&-&-&-&-&95.54&429.4$^{\dagger}$&&\\
AQM{+}{/}indA\cite{AQM}&-&-&-&-&94.64&516.1$^{\dagger}$&&\\ \cline{1-1}
\multirow{2}{*}{DST (Ours)}&6.25&2.59&8.31&13.14&\textbf{98.69}&\textbf{254.19}&& \\ \cline{2-9}
&8.73&3.55&10.49&15.96&98.76&184.65&v0.5 val&7,663\\
\hline
ADQ$^{\diamond}$\cite{ADQ}&-&-&-&-&94.99&400.8$^{\dagger}$&\multirow{4}{*}{v1.0 test}&\multirow{4}{*}{8,000}\\
RL{-}Q$^{\diamond}$\cite{RL}&-&-&-&-&93.38&529.6$^{\dagger}$&&\\
AMN\cite{AMN}&-&-&-&-&94.88&409.6$^{\dagger}$&& \\ 
DST (Ours)&33.49&17.47&30.62&33.97&\textbf{99.44}&\textbf{161.19}&& \\ \hline
SL{-}Q\cite{RL}&7.8&2.56&9.49&17.87&93.83&127.84&\multirow{5}{*}{v1.0 val}&\multirow{5}{*}{2,064}\\
ADQ\cite{ADQ}&10.73&3.39&14.82&25.29&95.73&87.92&\\
ReeQ-SL\cite{ReeQ}&31.21&17.78&45.01&59.98&99.00&20.60&&\\
ReeQ-RL\cite{ReeQ}&33.65&19.91&48.50&62.94&99.13&18.05&& \\ \cline{1-1}
\multirow{2}{*}{DST (Ours)}&\textbf{34.01}&\textbf{24.99}&\textbf{49.49}&\textbf{63.92}&\textbf{99.60}&\textbf{17.52}&&\\ \cline{2-9}
                                                   &4.10&2.27&5.88&8.13&98.02&1195.43&v0.9 val&40,504\\                                               
\hline
\end{tabular}}
\end{table*}

\begin{figure*}[!htb]
\centering
\includegraphics[width=1.0\textwidth,scale=0.5, clip=true,keepaspectratio]{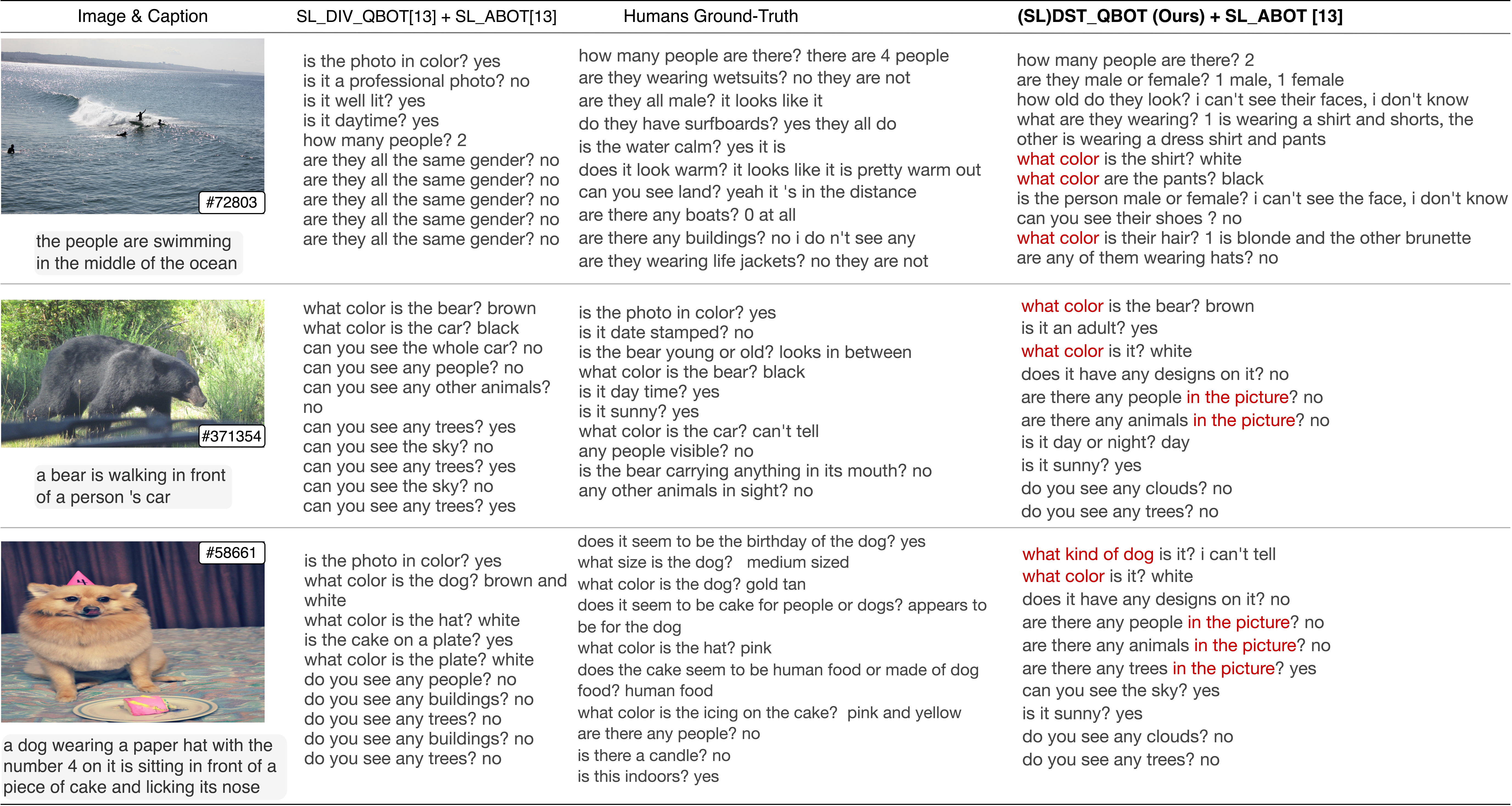}
\caption{Comparison of generated dialogs with ADQ \cite{ADQ} and humans on VisDial v1.0 val. Our QBot agent converses with SL{-}ABOT \cite{ADQ} for a fair comparison to ADQ model.}
\label{fig_cases}
\end{figure*}


\noindent {\bf Image Guessing}
The results of image guessing are provided in Table~\ref{tab_guess}.  Our DST model achieves significant improvements over previous state-of-the-art models (SOTA) across all metrics and datasets. On the validation split of v1.0, DST outperforms ReeQ and establishes new SOTA with a PMR of 99.60 and Mean of 17.52. On the test split of v1.0, DST consistently outperforms other strong models, such as AMN, with a PMR of 98.10. Compared to the real-image based SL-QI-G on the v0.5 test, DST achieves a PMR of 98.69, demonstrating the inefficiency of image retrieval from larger pools. DST also performs well on the v0.9 validation split with a PMR of 98.02. The trends in PMR are consistent across all datasets, and only DST shows a continuous increase in PMR as the dialogue progresses, highlighting its robustness and effectiveness in different dataset settings.

\begin{table}[!htb]
\centering
\caption{Ablation studies on major modules by removing it from the full model on VisDial v1.0 val.} \label{tab_abl}
\setlength{\tabcolsep}{1.8mm}{
\begin{tabular}{ll|ccccc|c}
\hline
\#&Model{\ }& MRR$\uparrow$& R@1$\uparrow$& R@5$\uparrow$& R@10$\uparrow$& Mean$\downarrow$& PMR$\uparrow$\\
\hline
1&Full model&\textbf{34.01}&\textbf{24.99}&\textbf{49.49}&\textbf{63.92}&\textbf{17.52}&\textbf{99.60} \\ \hline
2&$\quad$-VRDS &27.01&20.25&40.35&52.57&55.24&97.29\\ \hline
3&$\quad$-STrack &25.45&19.08&38.47&49.82&75.80&96.40\\ \hline
4&$\quad$-$\mathcal{L}_{\mathrm{MSE}}$&31.59&23.09&46.53&60.45&29.30&98.55\\ \hline
5&$\quad$-$\mathcal{L}_{\mathrm{PL}}$&32.28&23.61&47.83&62.16&25.30&99.11\\ \hline
\end{tabular}}
\end{table}

\noindent {\bf Ablation Studies}
Ablation studies on the v1.0 validation set (Table~\ref{tab_abl}) were performed to analyze the impact of each module. The full model's performance is reported in row 1, while subsequent rows correspond to the removal of specific modules to assess their significance.

Removing the VRDS module resulted in a significant drop in PMR by 2.30 points, an increase in Mean by 37.72 absolute points to 55.24, and a decrease in MRR by approximately 7 points to 27.01. These findings highlight the importance of three-hop reasoning using the recursive self-reference equation (R-SRE) for accurate image guessing. The role of the R-SRE operation is indirectly verified as it enables visually related reasoning, providing more distinguishing clues for generating visually related questions (refer to Fig.\ref{fig_cases} for details).

Removing the STrack module, which relies solely on the image caption at the 0th round, resulted in a decrease in PMR to 96.40, emphasizing the importance of caption information. The performance metrics of this configuration were lower than those of removing VRDS, indicating that the STrack module facilitates the incorporation of additional textual semantics and visual concepts into dialogue states. Additionally, the results from rows 4 and 5 demonstrate the effectiveness of both supervisions. Removing $\mathcal{L}_{\mathrm{PL}}$ led to a slight reduction in PMR, while removing $\mathcal{L}_{\mathrm{MSE}}$ decreased PMR by nearly 1 point, highlighting the efficacy of the combined $\mathcal{L}_{\mathrm{PL}}$ and $\mathcal{L}_{\mathrm{MSE}}$ in training the model efficiently.

\noindent {\bf Case Studies}
\noindent In Fig.\ref{fig_cases}, a comparison with recent ADQ \cite{ADQ} reveals two key observations. Firstly, our model effectively avoids repetition over the 10 rounds by combining VRDS and STrack modules. This prevents repetitive context and ensures that context vector $z^{(t)}$ remains distinctive and informative, leading to non-repetitive questions generated. Secondly, our model generates a higher number of visually related questions. In the first example, it asks three color-related questions (highlighted in red), while \cite{ADQ} and humans ask questions that are not focused on color. In the last two examples, our model initiates the dialogue with four and five image-related questions, respectively. These findings indicate that mental model of the unseen image enables the generation of image-like representations, prompting QBot to ask visually related questions.


\section{Conclusion}

\noindent This paper proposes DST, a novel dialogue state tracking approach for visual dialog question generation in the GuessWhich game. DST maintains and updates dialogue states, including word and mental image representations, enabling mentally related reasoning. Unlike previous studies, DST performs visual reasoning using mental representations of unseen images, achieving state-of-the-art performance. Future work will focus on exploring and visualizing the image state in DST.

\section*{Acknowledgements}
We thank the reviewers for their comments and suggestions. This paper was partially supported by the National Natural Science Foundation of China (NSFC 62076032), Huawei Noah’s Ark Lab, MoECMCC “Artificial Intelligence” Project (No. MCM20190701), Beijing Natural Science Foundation (Grant No. 4204100), and BUPT Excellent Ph.D. Students Foundation (No. CX2020309).
%
%
%
\bibliographystyle{splncs04}
\bibliography{bib_151}
\end{document}